\documentclass[10pt,twocolumn,letterpaper]{article}

\usepackage{booktabs}
\usepackage{tikz}
\usepackage{pgfplots}
\pgfplotsset{compat=1.18}
\usepackage{graphicx}
\usepackage{xcolor}
\usepackage{multirow}

\usepackage[pagenumbers]{cvpr} 

\definecolor{cvprblue}{rgb}{0.21,0.49,0.74}
\usepackage[pagebackref,breaklinks,colorlinks,allcolors=cvprblue]{hyperref}

\def\confName{CVPR}
\def\confYear{2027}

\title{Learning Where and How Much to Sample: Volume-Adaptive k-Space Acquisition for Accelerated MRI}

\author{
Reyhaneh Neshatavar$^{1,2}$
\qquad Mohsen Yavartanoo$^{1}$
\qquad Kyoung Mu Lee$^{1,3}$\\
$^{1}$IPAI, SNU, Seoul, Korea \quad
$^{2}$DGIST, Daegu, Korea \quad
$^{3}$Dept. of ECE \& ASRI, SNU, Seoul, Korea\\
{\tt\small \{reyhanehneshat,myavartanoo,kyoungmu\}@snu.ac.kr}
}

\begin{document}
\maketitle
\begin{abstract}
Magnetic resonance imaging (MRI) is an essential medical imaging technique, but its high costs and long acquisition times can limit accessibility.
Accelerated MRI reduces acquisition time through k-space undersampling, yet conventional methods often rely on predefined sampling patterns and uniform slice budgets, overlooking anatomical variations and inter-slice relationships.
In this paper, we propose a volume-adaptive k-space acquisition framework that jointly learns \emph{where} and \emph{how much} to sample across an entire MRI volume.
Using only acquired auto-calibration signal (ACS) measurements, our method exploits volume-wide anatomical information to adaptively allocate sampling budgets across slices while identifying informative k-space lines.
A learned joint slice--frequency refinement module coordinates sampling decisions while preserving an exact total volume-level acquisition budget.
To reconstruct the acquired measurements, we design a sensitivity-aware synchronous cross-slice reconstruction network that estimates coil sensitivities from ACS measurements and progressively refines neighboring-slice reconstructions while enforcing hard k-space data consistency.
Both networks are jointly optimized using reconstruction-based supervision without requiring fully sampled k-space at inference.
Experiments on knee and brain MRI across multiple contrasts and acceleration factors demonstrate the adaptive behavior of our acquisition framework and evaluate its reconstruction performance.
\end{abstract}
    
\section{Introduction}
\label{sec:intro}
Magnetic resonance imaging (MRI) provides detailed anatomical visualization without ionizing radiation, but its high costs and long acquisition times limit accessibility, reduce patient throughput, and increase sensitivity to motion.
Accelerated MRI addresses these limitations by acquiring fewer k-space measurements while maintaining reconstruction quality.
Recent deep learning methods have improved reconstruction through learned image priors and unrolled optimization, with approaches such as PromptMR+~\cite{xin2024rethinking} further exploiting neighboring-slice information.
However, many reconstruction methods rely on predefined sampling patterns, leaving acquisition largely independent of patient-specific anatomy.
Conventional uniform, random, or variable-density sampling typically assigns similar acquisition budgets across slices, despite variations in anatomical complexity and spatial frequency characteristics.
Furthermore, anatomical correlations between neighboring slices could support coordinated acquisition across the volume.
These observations motivate a fundamental question: \emph{where and how much should we sample in each slice to maximize reconstruction quality under a fixed volume-wide acquisition budget?}
Recent studies have explored adaptive k-space acquisition beyond predefined sampling patterns.
SUNO~\cite{gautam2026scan} selects scan-specific masks optimized offline using low-frequency measurements, while Dhar et al.~\cite{dhar2025learning} predict adaptive masks using pre-optimized patterns as supervision.
Himmetoglu and Konukoglu~\cite{HimMev_InterSlice_MICCAISAT2026} investigated reinforcement learning-based inter-slice budget allocation, while Yiasemis et al.~\cite{yiasemis2026end} jointly optimized adaptive sampling and reconstruction for dynamic MRI.
Nevertheless, jointly determining slice-specific acquisition budgets and frequency locations across an entire volume while exploiting inter-slice relationships under an exact acquisition budget remains challenging.

In this paper, we propose a \emph{volume-adaptive k-space acquisition framework} that jointly learns \emph{where} and \emph{how much} to sample across an MRI volume.
Our acquisition network employs a full-volume 3D CNN to extract anatomical features and inter-slice relationships from initially acquired auto-calibration signal (ACS) measurements.
A learned budget head determines slice-specific acquisition budgets, while an implicit coordinate-conditioned network predicts informative k-space line scores.
A differentiable soft sampling plan guides joint slice--frequency refinement before the final hard selection.
Conditioned on patient-specific anatomy, MRI contrast, and acceleration factor, the network coordinates sampling decisions under an exact volume-wide acquisition budget.
To reconstruct the acquired measurements, we develop a sensitivity-aware synchronous cross-slice reconstruction network that estimates complex coil-sensitivity maps from ACS measurements.
Multi-scale reconstruction cascades exploit neighboring-slice information and sampling masks, synchronously updating all slices while enforcing hard data consistency for each physical coil.
The acquisition and reconstruction networks are jointly optimized using reconstruction-based supervision.
At inference, the scanner first acquires central ACS lines from all slices, allowing the acquisition network to determine additional slice-specific sampling locations and budgets.
The selected measurements are then acquired and reconstructed without requiring fully sampled k-space or reference images.
We evaluate our framework on fastMRI multi-coil Knee~\cite{zbontar2018fastmri} and M4Raw low-field Brain~\cite{lyu2023m4raw}, covering different anatomies, MRI contrasts, and acceleration factors.
Our experiments demonstrate superior reconstruction quality under matched acquisition budgets, while achieving comparable performance at higher acceleration factors in certain settings.
The contributions of this work are summarized as follows:
\begin{itemize}
    \item We propose a patient-specific, contrast- and acceleration-aware acquisition network that jointly learns \emph{where} and \emph{how much} to sample in each slice under an exact volume-wide acquisition budget.
    \item We develop a sensitivity-aware synchronous cross-slice reconstruction network that exploits neighboring-slice information while enforcing hard multi-coil k-space data consistency.
    \item We demonstrate superior reconstruction performance on knee and brain MRI across multiple contrasts and acceleration factors, achieving comparable quality with fewer acquired k-space measurements in certain settings.
\end{itemize}
\section{Related Work}
\label{sec:related_work}

\begin{figure*}[t]
    \centering
    \includegraphics[width=0.98\textwidth]{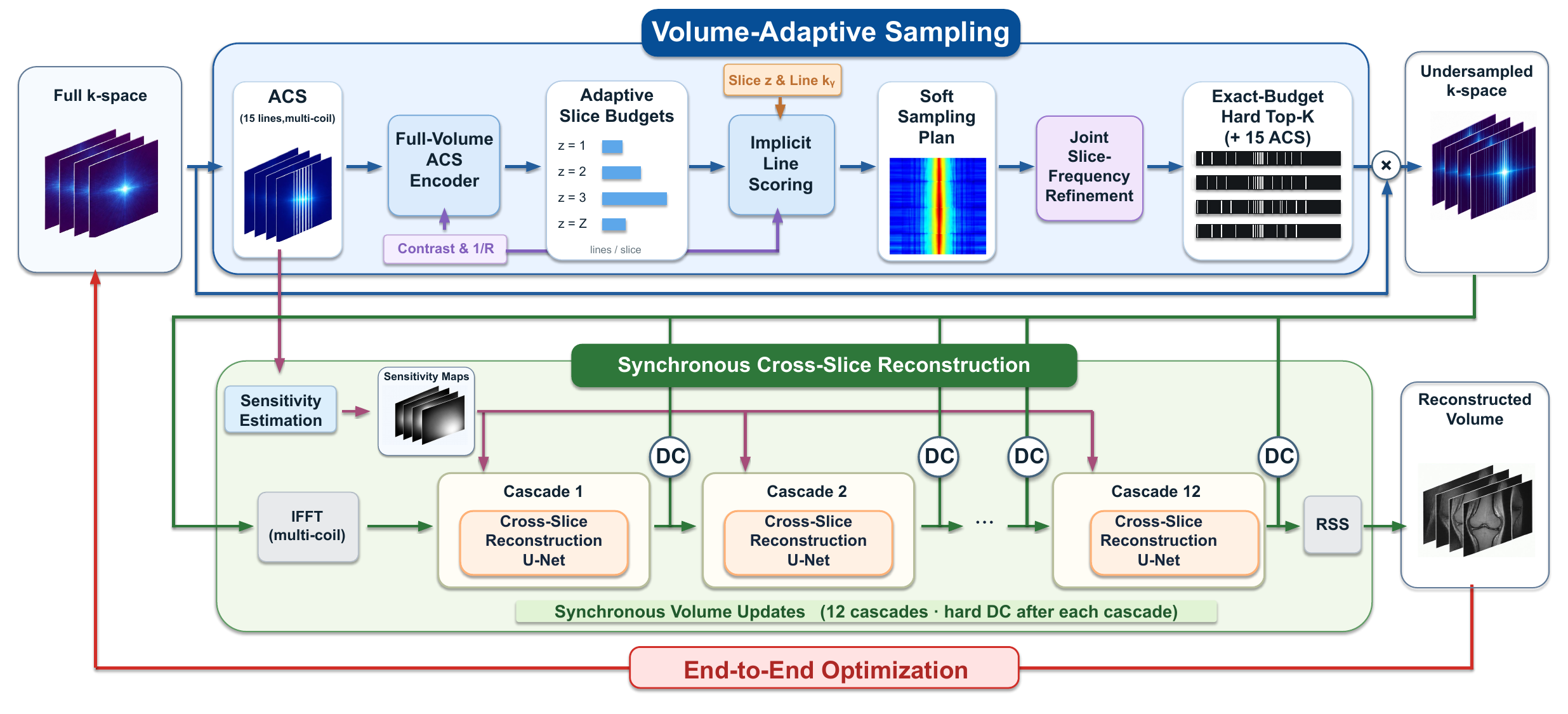}
    \caption{
    Overview of the proposed volume-adaptive MRI framework.
    A full-volume ACS encoder extracts anatomical features to predict adaptive slice budgets and informative k-space line scores.
    A differentiable soft sampling plan guides joint slice--frequency refinement before exact-budget hard selection.
    The acquired measurements are reconstructed through sensitivity-aware synchronous cross-slice reconstruction cascades with hard multi-coil k-space data consistency.
    Both networks are jointly optimized end-to-end.
    }
    \label{fig:overview}
\end{figure*}

\subsection{Learning-Based k-Space Sampling}

Conventional accelerated MRI~\cite{zbontar2018fastmri} employs predefined Cartesian undersampling patterns, such as equispaced or variable-density random masks, typically preserving central k-space measurements for calibration.
Although effective, these strategies do not explicitly adapt acquisition decisions to the anatomical and spectral characteristics of individual scans.
To address this limitation, LOUPE~\cite{bahadir2020deep} introduced a learning-based framework that jointly optimizes probabilistic sampling patterns and image reconstruction under a prescribed sampling budget.
J-MoDL~\cite{aggarwal2020j} further investigated joint optimization of sampling locations and model-based reconstruction using a continuous sampling formulation.
AutoSamp~\cite{alkan2024autosamp} employed variational information maximization to jointly optimize 3D MRI acquisition and reconstruction, outperforming conventional sampling strategies.
While these approaches establish the importance of reconstruction-aware sampling optimization, they primarily learn acquisition patterns for an imaging population or protocol rather than jointly coordinating slice-dependent sampling budgets across an entire volume.

\subsection{Scan-Adaptive MRI Acquisition}

Recent studies have explored scan-adaptive k-space sampling.
Pineda et al.~\cite{pineda2020active} formulated active MRI acquisition as a sequential decision process, selecting k-space measurements using reconstruction-related feedback.
Bakker et al.~\cite{bakker2022learning} extended adaptive acquisition to multi-coil MRI by learning sampling policies together with a reconstruction network.
Huang and Ravishankar~\cite{huang2022single} proposed MNet, which predicts object-specific sampling patterns from initially acquired low-frequency k-space measurements.

SUNO~\cite{gautam2026scan} optimized scan-specific Cartesian masks offline through coordinate descent and selected masks at inference by matching low-frequency measurements.
Dhar et al.~\cite{dhar2025learning} proposed a supervised scan-adaptive sampling framework that predicts acquisition patterns using pre-optimized masks as supervision.
Unlike these scan-level approaches, our framework jointly determines slice-specific acquisition budgets and frequency locations from full-volume ACS information under an exact acquisition-budget constraint.
Yiasemis et al.~\cite{yiasemis2026end} jointly optimized adaptive sampling and reconstruction across dynamic MRI frames, rather than coordinating a shared acquisition budget across anatomical slices.
Although these approaches demonstrate the benefits of adaptive frequency selection, temporal sampling coordination differs from distributing a shared acquisition budget across neighboring anatomical slices.

\subsection{Joint Inter-Slice Acquisition}

Most scan-adaptive sampling approaches emphasize the selection of informative k-space measurements for individual scans or slices.
However, neighboring MRI slices exhibit anatomical correlations that can be exploited when planning acquisitions across an entire volume.
Himmetoglu and Konukoglu~\cite{HimMev_InterSlice_MICCAISAT2026} investigated nonuniform inter-slice acquisition budgets to prioritize pathology-relevant regions.
Their method sequentially allocates fixed-size groups of k-space lines to selected slices, with frequency locations chosen randomly.
In contrast, our framework jointly learns the number and frequency locations of acquired lines for each slice, with single-line allocation precision and an exact volume-wide acquisition budget.
Complementary sampling across slices has also been explored for MRI reconstruction.
Anvari-Fard and Soltanian-Zadeh~\cite{anvari2026reconstruction} employed alternating complementary undersampling masks across adjacent slices in dual-contrast brain MRI, exploiting neighboring measurements to estimate missing k-space information.
Unlike their predefined complementary masks, our framework jointly learns slice-dependent sampling budgets and frequency locations from full-volume ACS measurements.
This enables coordinated acquisition across slices without imposing predefined complementary patterns.

\subsection{Multi-Slice MRI Reconstruction}

Reconstruction networks have improved image recovery from undersampled multi-coil k-space measurements.
PromptMR+~\cite{xin2024rethinking} enhanced unrolled reconstruction through information flow and the use of adjacent-slice information for sensitivity estimation and reconstruction.
More recently, MR-DiffRecon~\cite{AlWaf_MRDiffRecon_MICCAISAT2026} incorporated multi-scale adjacent-slice feature alignment and implicit prompting to improve reconstruction efficiency across different MRI settings.
Unlike reconstruction-focused approaches that operate on prescribed sampling masks, our framework jointly learns volume-adaptive acquisition and sensitivity-aware reconstruction.
It further enables synchronous information exchange across slices throughout iterative reconstruction.
\section{Method}
\label{sec:method}

As illustrated in Fig.~\ref{fig:overview}, our framework jointly learns \emph{where} and \emph{how much} to sample across an MRI volume using initially acquired auto-calibration signal (ACS) measurements and known acquisition settings.

\subsection{Problem Formulation}
\label{sec:problem}
Consider a multi-coil MRI volume with $Z$ physical slices.
Let $\mathbf{Y}_z\in\mathbb{C}^{C\times H\times W}$ denote the fully sampled k-space of slice $z$, where $C$ is the number of receiver coils, $H$ is the readout dimension, and $W$ is the phase-encoding dimension.
For Cartesian acquisition, a binary mask $\mathbf{m}_z\in\{0,1\}^{W}$ specifies the acquired phase-encoding lines.
The corresponding undersampled measurements are
\begin{equation}
    \widetilde{\mathbf{Y}}_z
    =
    \mathbf{m}_z\odot\mathbf{Y}_z,
    \label{eq:masked_kspace}
\end{equation}
where $\mathbf{m}_z$ is broadcast over the receiver coils and the entire readout dimension.
Thus, selecting one mask entry acquires one complete readout line, rather than an isolated k-space point.
Let $\Omega_z\subseteq\{1,\ldots,W\}$ denote the valid phase-encoding support of slice $z$; padded or unsupported positions are excluded from selection.
The number of acquired lines in slice $z$ is $K_z=\|\mathbf{m}_z\|_1$.
Given a prescribed integer budget $B$ for the entire volume, we seek to jointly determine
\begin{equation}
    \underbrace{K_z}_{\text{How much}}
    \qquad\text{and}\qquad
    \underbrace{\mathbf{m}_z}_{\text{Where}},
    \quad\text{s.t.}\quad
    \sum_{z=1}^{Z}K_z=B.
    \label{eq:where_howmuch}
\end{equation}

\subsection{Volume-Adaptive Sampling}
\label{sec:sampling}

\paragraph{ACS-Guided Volume Feature Extraction.}
\label{sec:acs}
Acquisition decisions must be made without observing unmeasured k-space.
We define a fixed central ACS mask $\mathbf{a}\in\{0,1\}^{W}$, containing $A=\|\mathbf{a}\|_1=15$ phase-encoding lines per slice.
The initially available data are
\begin{equation}
    \mathbf{Y}_z^{\mathrm{ACS}}
    =
    \mathbf{a}\odot\mathbf{Y}_z,
    \qquad
    \mathcal{A}
    =
    \{\mathbf{Y}_z^{\mathrm{ACS}}\}_{z=1}^{Z}.
    \label{eq:volume_acs}
\end{equation}
To capture anatomical and spectral information across the entire volume, we employ an ACS-conditioned 3D convolutional encoder.
The acquired complex ACS measurements are transformed along the fully sampled readout direction, and their real, imaginary, and magnitude components are used as network inputs.
All physical slices are processed jointly at the native readout resolution using a shared 3D CNN, while features from different receiver coils are aggregated to obtain coil-independent representations.
The resulting features are further processed by dilated inter-slice convolutions to capture anatomical relationships across the volume.
Slice-specific and global features are combined into volume-aware representations $\mathbf{F}=f_{\mathrm{vol}}(\mathcal{A})$, which guide adaptive budget allocation and line selection.
This enables acquisition decisions to exploit information from the entire MRI volume without requiring unacquired k-space, as illustrated in Fig.~\ref{fig:overview}.

\begin{table*}[t]
\centering
\caption{Quantitative comparison on the fastMRI multi-coil
knee dataset at $4\times$ and $8\times$ acceleration.
PSNR (dB) and SSIM (\%) are reported as mean
$\pm$ standard deviation.
Published baseline values are taken from
PromptMR+~\cite{xin2024rethinking}.
Dashes indicate unavailable results.}
\label{tab:knee_results}
\scriptsize
\setlength{\tabcolsep}{2pt}
\renewcommand{\arraystretch}{1.15}
\begin{tabular*}{\textwidth}{@{\extracolsep{\fill}}l*{8}{c}}
\toprule
& \multicolumn{4}{c}{$4\times$ Acceleration}
& \multicolumn{4}{c}{$8\times$ Acceleration} \\
\cmidrule(lr){2-5}\cmidrule(lr){6-9}
& \multicolumn{2}{c}{PD}
& \multicolumn{2}{c}{PDFS}
& \multicolumn{2}{c}{PD}
& \multicolumn{2}{c}{PDFS} \\
\cmidrule(lr){2-3}\cmidrule(lr){4-5}
\cmidrule(lr){6-7}\cmidrule(lr){8-9}
Method
& PSNR$\uparrow$ & SSIM$\uparrow$
& PSNR$\uparrow$ & SSIM$\uparrow$
& PSNR$\uparrow$ & SSIM$\uparrow$
& PSNR$\uparrow$ & SSIM$\uparrow$ \\
\midrule
Inter-Slice~\cite{HimMev_InterSlice_MICCAISAT2026}
& $35.23\pm1.54$ & $89.36\pm2.64$
& $35.14\pm3.11$ & $85.95\pm4.79$
& $33.88\pm2.20$ & $87.06\pm3.21$
& $34.46\pm1.58$ & $82.67\pm4.75$ \\

U-Net~\cite{ronneberger2015u}
& $37.27\pm1.76$ & $92.11\pm2.72$
& $37.07\pm2.47$ & $88.03\pm5.04$
& $35.41\pm2.14$ & $90.36\pm3.13$
& $34.84\pm1.59$ & $83.71\pm5.04$ \\

i-RIM~\cite{putzky2019invert}
& $39.61\pm2.33$ & $94.16\pm2.87$
& $38.23\pm3.13$ & $89.27\pm4.96$
& $37.91\pm2.47$ & $93.22\pm2.85$
& $35.84\pm1.71$ & $85.27\pm4.93$ \\

E2E-VarNet~\cite{sriram2020end}
& $40.09\pm2.32$ & $94.56\pm2.64$
& $38.43\pm3.24$ & $89.50\pm4.95$
& $38.72\pm2.50$ & $93.93\pm2.68$
& $36.16\pm1.76$ & $85.70\pm4.94$ \\

HUMUS-Net~\cite{fabian2022humus}
& -- & -- & -- & --
& $38.72\pm2.50$ & $94.07\pm2.68$
& $35.98\pm1.75$ & $85.77\pm4.93$ \\

HUMUS-Net-L
& -- & -- & -- & --
& $39.03\pm2.47$ & $94.18\pm2.61$
& $36.19\pm1.79$ & $85.84\pm4.92$ \\

PromptMR~\cite{xin2023fill}
& $40.58\pm2.59$ & $94.87\pm2.75$
& $38.59\pm3.33$ & $89.75\pm4.94$
& $39.48\pm2.54$ & $94.51\pm2.54$
& $36.42\pm1.79$ & $86.10\pm4.93$ \\

PromptMR+~\cite{xin2024rethinking}
& $40.85\pm2.74$ & $95.00\pm2.84$
& $38.71\pm3.37$ & $89.85\pm4.92$
& $40.09\pm2.58$ & $94.90\pm2.45$
& $36.49\pm1.88$ & $86.29\pm4.95$ \\

\midrule
Ours
& $\textbf{41.16}\pm\textbf{2.73}$ & $\textbf{95.21}\pm\textbf{2.72}$ & $\textbf{38.86}\pm\textbf{3.33}$ & $\textbf{89.94}\pm\textbf{4.65}$
& $\textbf{40.85}\pm\textbf{2.24}$& $\textbf{95.44}\pm\textbf{2.15}$
& $\textbf{36.96}\pm\textbf{1.86}$& $\textbf{86.75}\pm\textbf{4.93}$\\
\bottomrule
\end{tabular*}
\end{table*}

\begin{table*}[t]
\centering
\caption{
Quantitative comparison on the M4Raw low-field brain MRI
dataset at $4\times$ and $8\times$ acceleration.
}
\label{tab:m4raw_results}

\scriptsize
\setlength{\tabcolsep}{1pt}
\renewcommand{\arraystretch}{1.15}

\begin{tabular*}{\textwidth}{
@{\extracolsep{\fill}}l*{12}{c}
}
\toprule
& \multicolumn{6}{c}{$4\times$ Acceleration}
& \multicolumn{6}{c}{$8\times$ Acceleration} \\
\cmidrule(lr){2-7}
\cmidrule(lr){8-13}

& \multicolumn{2}{c}{T1}
& \multicolumn{2}{c}{T2}
& \multicolumn{2}{c}{FLAIR}
& \multicolumn{2}{c}{T1}
& \multicolumn{2}{c}{T2}
& \multicolumn{2}{c}{FLAIR} \\
\cmidrule(lr){2-3}
\cmidrule(lr){4-5}
\cmidrule(lr){6-7}
\cmidrule(lr){8-9}
\cmidrule(lr){10-11}
\cmidrule(lr){12-13}

Method
& PSNR$\uparrow$ & SSIM$\uparrow$
& PSNR$\uparrow$ & SSIM$\uparrow$
& PSNR$\uparrow$ & SSIM$\uparrow$
& PSNR$\uparrow$ & SSIM$\uparrow$
& PSNR$\uparrow$ & SSIM$\uparrow$
& PSNR$\uparrow$ & SSIM$\uparrow$ \\
\midrule

E2E-VarNet~\cite{sriram2020end}
& 33.13 & 82.64 & 30.83 & 76.02 & 31.10 & 76.77
& 30.98 & 78.69 & 28.65 & 70.47 & 29.21 & 71.85 \\

MoDL~\cite{aggarwal2018modl}
& 31.97 & 80.79 & 30.02 & 73.70 & 30.15 & 74.56
& 29.39 & 75.69 & 27.29 & 66.27 & 27.98 & 68.63 \\

PromptMR~\cite{xin2023fill}
& 33.63 & 83.29 & 31.19 & 76.82 & 31.40 & 77.45
& 31.74 & 79.80 & 29.29 & 71.99 & 29.73 & 73.08\\


\midrule

\textbf{Ours}
& \textbf{33.85} & \textbf{83.79} & \textbf{31.90} & \textbf{78.04} & \textbf{32.29} & \textbf{78.65}
& \textbf{32.08} & \textbf{80.07} & \textbf{29.93} & \textbf{72.40} & \textbf{30.47} & \textbf{73.31} \\

\bottomrule
\end{tabular*}
\end{table*}

\paragraph{Adaptive Slice Budget Allocation.}
\label{sec:budget}
Uniform slice-wise acquisition budgets overlook differences in anatomical content and reconstruction requirements across a volume.
Our acquisition network instead predicts slice-specific sampling preferences using the volume-aware ACS representations, the known MRI contrast $c$, and the requested sampling fraction $r=1/R$.
For a requested acceleration factor $R$, the total acquisition budget is determined from the physically valid phase-encoding support:
\begin{equation}
    B
    =
    \operatorname{round}\left(
        \frac{1}{R}
        \sum_{z=1}^{Z}|\Omega_z|
    \right).
    \label{eq:nominal_budget}
\end{equation}
This defines an exact integer acquisition budget for each volume while accounting for its physical acquisition dimensions.
Let $n_z=K_z-A$ denote the number of additional, non-ACS lines allocated to slice $z$.
The allocation satisfies
\begin{equation}
\begin{gathered}
    \sum_{z=1}^{Z}n_z=B-AZ,\\
    0\le n_z\le|\Omega_z|-A,
    \qquad n_z\in\mathbb{Z}.
    \label{eq:volume_budget}
\end{gathered}
\end{equation}
A learned budget head predicts continuous allocation preferences from the volume features and acquisition conditions, which are converted to valid integer budgets with single-line precision and an exact volume-wide sum.
No fixed per-slice budget or additional percentage-based deviation bound is imposed beyond the ACS minimum and physically valid support.
During training, differentiable budget allocation enables gradient propagation to the budget predictor while preserving integer line counts in forward acquisition plan.

\paragraph{Contrast- and Acceleration-Aware Line Selection.}
\label{sec:contrast}
The locations of informative k-space lines can depend on patient-specific anatomy, MRI contrast, and the available sampling budget.
We therefore condition the line-scoring network on the volume-aware ACS features, the known contrast label $c$, the continuous sampling fraction $r=1/R$, and the learned slice budgets $\mathbf{K}=(K_1,\ldots,K_Z)$.
The predicted scores are
\begin{equation}
    \mathbf{S}
    =
    f_{\mathrm{score}}(\mathbf{F},c,r,\mathbf{K}),
    \qquad
    \mathbf{S}\in\mathbb{R}^{Z\times W},
    \label{eq:utility_score}
\end{equation}
where $S_{z,k}$ estimates the utility of acquiring phase-encoding line $k$ in slice $z$.
The scoring network employs an implicit multi-layer perceptron (MLP) that combines ACS-derived anatomical features with the normalized slice position, the native phase-encoding coordinate, and their Fourier-feature encodings.
The learned slice budget is also incorporated into the scoring process, allowing the selection preferences to adapt to the number of available measurements.
Contrast conditioning accounts for acquisition protocols such as PD and PDFS, while the sampling fraction $r$ informs the network of the requested acceleration.
By directly evaluating individual phase-encoding coordinates, the network predicts sampling scores without relying on predefined frequency patterns or interpolation from a fixed-resolution score map.

\paragraph{Joint Slice--Frequency Coordination and Hard Sampling.}
\label{sec:complementary}
Selecting k-space lines independently for each slice may overlook relationships between sampling decisions across the MRI volume.
To coordinate these decisions, we first construct a differentiable preliminary soft sampling plan from predicted line scores and allocated slice budgets:
\begin{equation}
    \mathbf{P}
    =
    \operatorname{SoftTopK}
    \left(
        \mathbf{S},\boldsymbol{\Gamma},\mathbf{n}
    \right),
    \qquad
    \mathbf{P}\in[0,1]^{Z\times W},
    \label{eq:soft_sampling}
\end{equation}
where $\boldsymbol{\Gamma}$ denotes the eligible non-ACS sampling locations and $\mathbf{n}=(n_1,\ldots,n_Z)$ represents the allocated additional-line budgets.
Each element $P_{z,k}$ indicates the continuous selection preference for phase-encoding line $k$ in slice $z$, with the soft selection constrained by the corresponding slice budget.
Unlike a fixed sampling mask, $\mathbf{P}$ is dynamically generated from the learned scores for each MRI volume, providing a differentiable representation of preliminary acquisition preferences.
To coordinate these preferences across slices, we introduce a learned joint slice--frequency 2D CNN that considers the preliminary sampling plan, neighboring-slice preferences, volume-wide sampling coverage, phase-encoding coordinates, and allocated slice budgets.
The module refines the initial scores through a learned residual correction:
\begin{equation}
    \widehat{\mathbf{S}}
    =
    \mathbf{S}
    +
    f_{\mathrm{joint}}
    \left(
        \mathbf{S},\mathbf{P},\mathbf{K}
    \right),
    \label{eq:volume_refinement}
\end{equation}
where the positional information and aggregated sampling preferences are incorporated within $f_{\mathrm{joint}}$.
This allows sampling decisions to be adjusted by considering acquisition preferences throughout the volume before any binary selection.
Following refinement, we perform a single hard selection to obtain the final acquisition masks:
\begin{equation}
    \begin{aligned}
    \Gamma_z
    &=
    \Omega_z\setminus\Omega_{\mathrm{ACS}},\\
    \mathbf{m}^{\mathrm{hard}}_z(k)
    &=
    \mathbf{a}(k)
    +
    \mathbf{1}\!\left[
        k\in
        \operatorname{TopK}_{n_z}
        \left(
            \widehat{\mathbf{S}}_{z,\Gamma_z}
        \right)
    \right].
    \end{aligned}
    \label{eq:hard_sampling}
\end{equation}
Here, $\Omega_{\mathrm{ACS}}$ denotes the mandatory calibration region and $\operatorname{TopK}_{n_z}$ returns the indices of the $n_z$ highest-scoring eligible lines.
By construction, the resulting masks retain all mandatory ACS measurements, exclude physically invalid locations, and satisfy the exact volume-wide acquisition budget:
\begin{equation}
    \sum_{z=1}^{Z}
    \left\|
        \mathbf{m}^{\mathrm{hard}}_z
    \right\|_1
    =
    B.
    \label{eq:exact_hard_budget}
\end{equation}
Importantly, the learned coordination allows neighboring slices to acquire different k-space locations without imposing predefined complementary patterns or explicitly enforcing non-overlapping measurements.
Measurements acquired in one physical slice are not treated as measurements of another; instead, the reconstruction network exploits anatomical correlations across slices to integrate information from their distinct acquired measurements.

\subsection{Synchronous Cross-Slice Reconstruction}
\label{sec:reconstruction}
\paragraph{Sensitivity Estimation.}
We estimate full-resolution complex coil-sensitivity maps using only the initially acquired ACS measurements.
For each slice, a lightweight multi-scale encoder--decoder processes the ACS-only complex coil images from five neighboring slices:
\begin{equation}
    \mathbf{s}_{z}
    =
    \operatorname{Norm}_{C}\!\left(
        g_{\phi}\left(
            \{\mathcal{F}^{-1}(\mathbf{Y}^{\mathrm{ACS}}_j)\}_{j\in\mathcal{N}(z)}
        \right)
    \right),
    \label{eq:acs_sensitivity}
\end{equation}
where $\operatorname{Norm}_{C}$ denotes coil-wise root-sum-of-squares normalization.
The estimated sensitivity maps are computed once from ACS measurements and reused throughout reconstruction.
%

\definecolor{mriI0}{RGB}{0,0,4}
\definecolor{mriI1}{RGB}{11,7,36}
\definecolor{mriI2}{RGB}{33,12,74}
\definecolor{mriI3}{RGB}{61,9,101}
\definecolor{mriI4}{RGB}{87,16,110}
\definecolor{mriI5}{RGB}{113,25,110}
\definecolor{mriI6}{RGB}{138,34,106}
\definecolor{mriI7}{RGB}{163,44,97}
\definecolor{mriI8}{RGB}{188,55,84}
\definecolor{mriI9}{RGB}{210,70,68}
\definecolor{mriI10}{RGB}{228,90,49}
\definecolor{mriI11}{RGB}{241,115,29}
\definecolor{mriI12}{RGB}{249,142,9}
\definecolor{mriI13}{RGB}{252,172,17}
\definecolor{mriI14}{RGB}{249,203,53}
\definecolor{mriI15}{RGB}{242,234,105}
\definecolor{mriI16}{RGB}{252,255,164}

\newcommand{\ErrorColorBar}{%
\begin{tikzpicture}[baseline=0pt]

  \foreach \low/\high in {
    0/1,1/2,2/3,3/4,4/5,5/6,6/7,7/8,
    8/9,9/10,10/11,11/12,12/13,
    13/14,14/15,15/16
  }{%
    \pgfmathsetmacro{\yb}{2.25*\low/16}
    \pgfmathsetmacro{\yt}{2.25*\high/16}

    \shade[
  bottom color/.expanded={mriI\low},
  top color/.expanded={mriI\high}
]
(0,\yb) rectangle (0.24,\yt);
  }

  \draw[thin] (0,0) rectangle (0.24,2.25);

  \foreach \pos/\lab in {
    0/0,
    0.45/1,
    0.90/2,
    1.35/3,
    1.80/4,
    2.25/5
  }{%
    \draw (0.24,\pos) -- (0.32,\pos);
    \node[
      anchor=west,
      font=\scriptsize
    ] at (0.36,\pos) {\lab};
  }

  \node[
    anchor=south,
    font=\tiny
  ] at (0.29,2.37) {Error (\%)};

\end{tikzpicture}%
}
\newcommand{\CenteredErrorImage}[1]{%
  $\vcenter{\hbox{%
    \includegraphics[width=0.145\textwidth]{#1}%
  }}$%
}

\newcommand{\CenteredErrorBar}{%
  $\vcenter{\hbox{\ErrorColorBar}}$%
}
\begin{figure*}[t]
\centering
\setlength{\tabcolsep}{3pt}
\renewcommand{\arraystretch}{1.05}
\scriptsize

\begin{tabular}{lcccccc}
\toprule
& \textbf{Inter-Slice}
& \textbf{i-RIM}
& \textbf{E2E-VarNet}
& \textbf{PromptMR+}
& \textbf{Ours}
& \textbf{Ground Truth} \\
\midrule


\multirow{2}{*}{\textbf{PD}}
&
\shortstack[c]{
\includegraphics[width=0.145\textwidth]{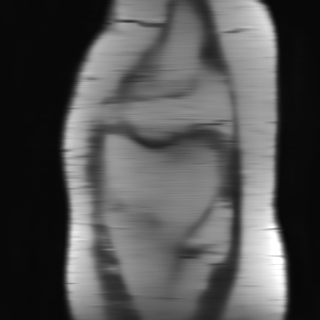}\\
29.38 dB / 80.44
}
&
\shortstack[c]{
\includegraphics[width=0.145\textwidth]{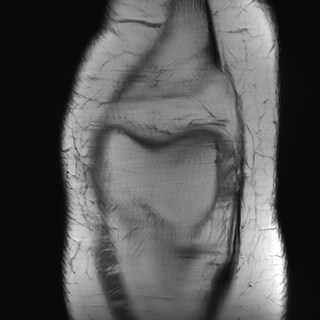}\\
34.73 dB / 90.49
}
&
\shortstack[c]{
\includegraphics[width=0.145\textwidth]{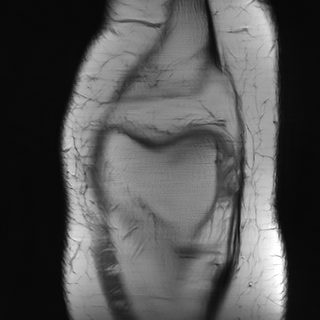}\\
35.41 dB / 91.31
}
&
\shortstack[c]{
\includegraphics[width=0.145\textwidth]{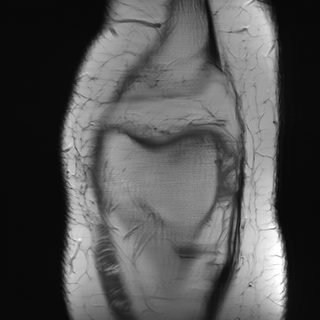}\\
37.22 dB / 93.50
}
&
\shortstack[c]{
\includegraphics[width=0.145\textwidth]{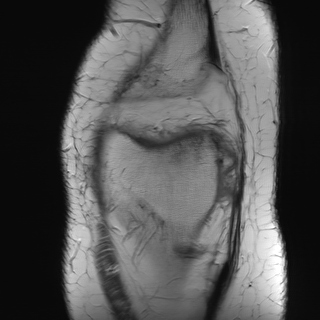}\\
\textbf{39.29 dB / 95.35}
}
&
\shortstack[c]{
\includegraphics[width=0.145\textwidth]{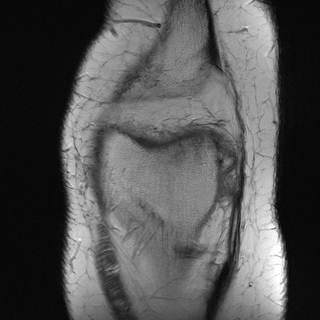}\\
\textbf{PSNR / SSIM}
}
\vspace{-3px}
\\ 
&
\CenteredErrorImage{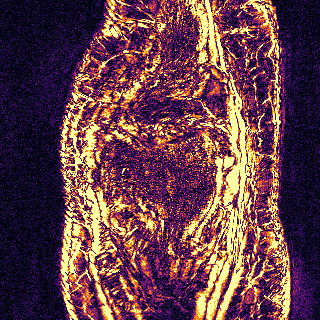}
&
\CenteredErrorImage{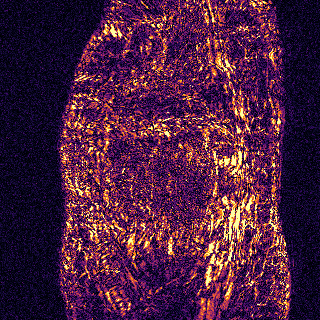}
&
\CenteredErrorImage{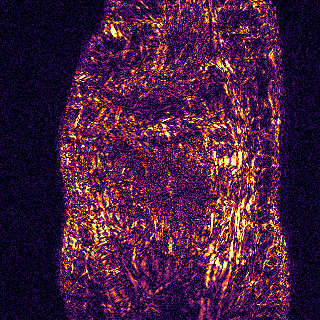}
&
\CenteredErrorImage{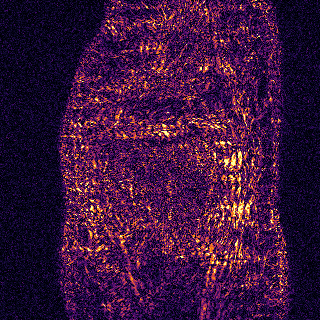}
&
\CenteredErrorImage{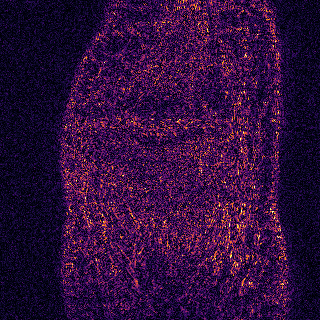}
&
\multicolumn{1}{@{}l}{\hspace{3pt}\CenteredErrorBar}
\\

\midrule


\multirow{2}{*}{\textbf{PDFS}}
&
\shortstack[c]{
\includegraphics[width=0.145\textwidth]{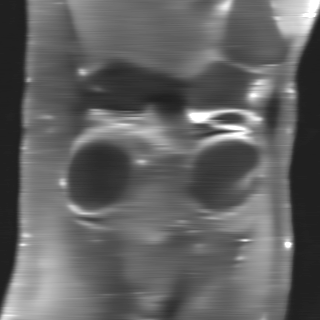}\\
30.41 dB / 74.67
}
&
\shortstack[c]{
\includegraphics[width=0.145\textwidth]{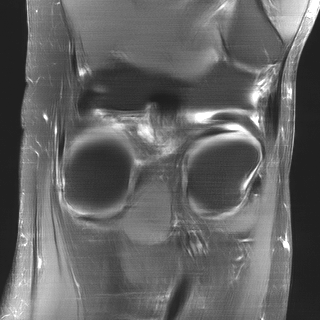}\\
33.68 dB / 84.33
}
&
\shortstack[c]{
\includegraphics[width=0.145\textwidth]{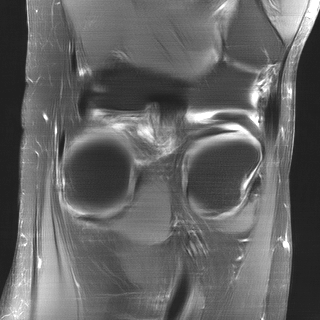}\\
34.08 dB / 85.27 
}
&
\shortstack[c]{
\includegraphics[width=0.145\textwidth]{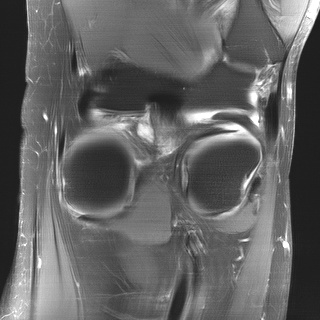}\\
34.90 dB / 86.74
}
&
\shortstack[c]{
\includegraphics[width=0.145\textwidth]{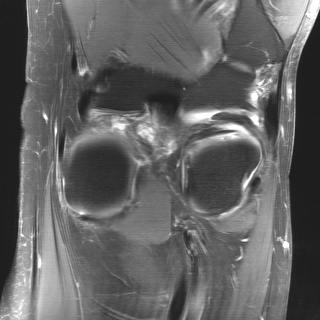}\\
\textbf{36.57 dB / 89.02}
}
&
\shortstack[c]{
\includegraphics[width=0.145\textwidth]{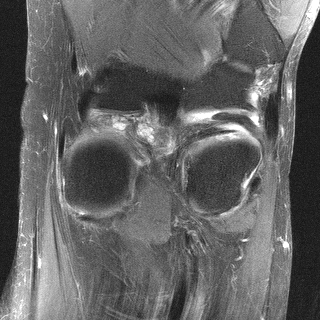}\\
\textbf{PSNR / SSIM}
}
\vspace{-3px}
\\
&
\CenteredErrorImage{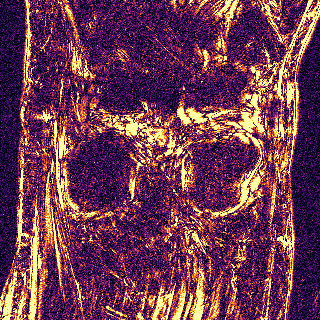}
&
\CenteredErrorImage{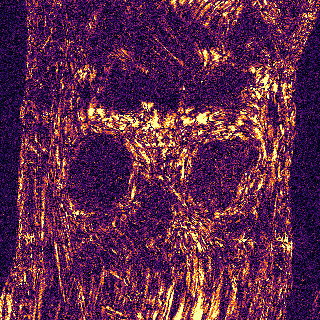}
&
\CenteredErrorImage{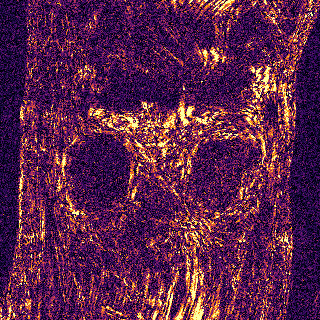}
&
\CenteredErrorImage{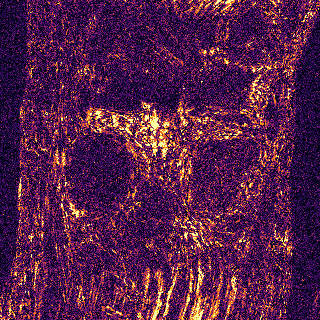}
&
\CenteredErrorImage{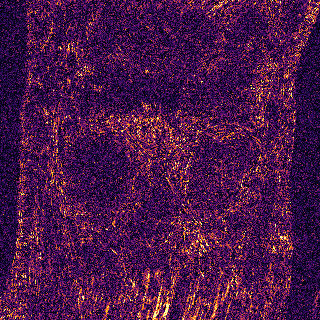}
&
\multicolumn{1}{@{}l}{\hspace{3pt}\CenteredErrorBar}
\\

\bottomrule
\end{tabular}

\caption{
Qualitative comparison on fastMRI Knee at $8\times$ acceleration.
The first and third rows show PD and PDFS reconstructions, while the second and fourth rows show their absolute error maps against fully sampled ground truth.}
\label{fig:knee_qualitative}
\end{figure*}

\paragraph{Acquisition-Aware Reconstruction.}
The acquired multi-coil k-space measurements are transformed into complex coil images to initialize reconstruction.
At each cascade, estimated sensitivity maps combine the coil images into a complex image representation, which a multi-scale U-Net refines using neighboring-slice information and sampling masks.
The network is conditioned on MRI contrast and sampling fraction $r=1/R$ through learned embeddings, while recurrent features share information across cascades.
Each cascade predicts a two-channel complex-image residual, which is expanded to individual coils using the sensitivity maps before data consistency.

\paragraph{Synchronous Updates and Data Consistency.}
At each cascade, all slices are reconstructed using the same current volume state, and their refined estimates are shared synchronously with neighboring slices in the following cascade.
Let $\mathbf{Q}_{z,c}^{(t)}$ denote the complex k-space estimate of coil $c$ in slice $z$ at cascade $t$, initialized as $\mathbf{Q}_{z,c}^{(0)}=\widetilde{\mathbf{Y}}_{z,c}$.
The corresponding coil image is $\mathbf{u}_{z,c}^{(t)}=\mathcal{F}^{-1}(\mathbf{Q}_{z,c}^{(t)})$.
At each cascade, the network predicts a complex-image residual $\Delta\mathbf{v}^{(t)}_z$, which is mapped to each physical coil using its estimated sensitivity map $\mathbf{s}_{z,c}$.
The updated k-space estimate is obtained by applying hard data consistency after each cascade:
\begin{equation}
\begin{aligned}
\mathbf{Q}_{z,c}^{(t+1)}
={}&
(1-\mathbf{m}_z^{\mathrm{hard}})
\odot\mathcal{F}\!\left[
    \mathbf{u}_{z,c}^{(t)}
    +
    \mathbf{s}_{z,c}\odot\Delta\mathbf{v}^{(t)}_z
\right]\\
&+
\mathbf{m}_z^{\mathrm{hard}}
\odot\widetilde{\mathbf{Y}}_{z,c}.
\end{aligned}
\label{eq:hard_dc}
\end{equation}
The sampling mask is broadcast over the readout dimension, and data consistency is enforced independently for every physical coil.
This preserves all acquired k-space measurements while allowing the network to estimate unacquired coefficients.
After the final cascade, the reconstructed coil images are combined using root-sum-of-squares to obtain the final magnitude MRI volume.

\subsection{Training and Inference}
\label{sec:train_inference}

\paragraph{Joint Optimization.}
\label{sec:training}
The acquisition network, sensitivity estimator, and reconstruction network are jointly optimized end-to-end using reconstruction-based supervision.
The training objective is applied only to the final reconstruction:
\begin{equation}
    \mathcal{L}_{\mathrm{rec}}
    =
    \mathcal{L}_{\mathrm{SSIM}}
    +
    100\,\mathcal{L}_{\mathrm{img}}
    +
    0.01\,\mathcal{L}_{k},
    \label{eq:reconstruction_loss}
\end{equation}
where $\mathcal{L}_{\mathrm{SSIM}}$ compares the reconstructed and reference magnitude images, $\mathcal{L}_{\mathrm{img}}$ is the mean squared image error normalized by the squared reference intensity range, and $\mathcal{L}_{k}$ combines global and frequency-band-normalized complex multi-coil k-space errors at physically valid unacquired locations.
No intermediate-cascade reconstruction losses are applied.
To enable gradient propagation through discrete sampling decisions, we apply a straight-through estimator:
\begin{equation}
    \mathbf{m}^{\mathrm{ST}}_z
    =
    \mathbf{m}^{\mathrm{hard}}_z
    +
    \boldsymbol{\pi}_z
    -
    \operatorname{sg}(\boldsymbol{\pi}_z),
    \label{eq:straight_through}
\end{equation}
where $\boldsymbol{\pi}_z$ is a differentiable soft selection derived from the refined line scores and slice budget, and $\operatorname{sg}$ denotes the stop-gradient operator.
This preserves exact binary sampling in the forward pass while allowing gradients to update the acquisition policy.
Fully sampled k-space and reference images are used only for training supervision and are not provided as acquisition-network inputs.

\paragraph{Inference.}
\label{sec:inference}
At inference, the scanner first acquires central ACS lines from all slices.
Using these measurements and known acquisition settings, the sampling network selects additional k-space lines under the exact volume-wide budget.
The acquired multi-coil measurements are reconstructed using ACS-derived coil sensitivities and synchronous cross-slice reconstruction, without requiring fully sampled k-space or reference images.

\section{Experiments}
\label{sec:experiments}

\subsection{Experimental Setup}
\label{sec:experimental_setup}

\paragraph{Datasets.}
We evaluate on fastMRI multi-coil Knee~\cite{zbontar2018fastmri} (PD/PDFS) and M4Raw low-field Brain~\cite{lyu2023m4raw} (T1/T2/FLAIR).
For Knee, we use 973 training volumes and split the 199 official validation volumes into 99 for validation and 100 for testing, following previous works.
For M4Raw, we use subject-disjoint splits of 128 training and 30 validation subjects, evaluating only the first acquisition repetition per subject and contrast.
Following PromptMR+~\cite{xin2024rethinking}, Knee reconstructions and provided RSS references are center-cropped to $320\times320$, while M4Raw RSS references are computed without averaging acquisition repetitions.
Knee acceleration categories follow official sampling metadata, while each M4Raw validation volume is evaluated at both acceleration factors.

\begin{figure*}[t]
    \centering
    \includegraphics[width=0.98\textwidth]{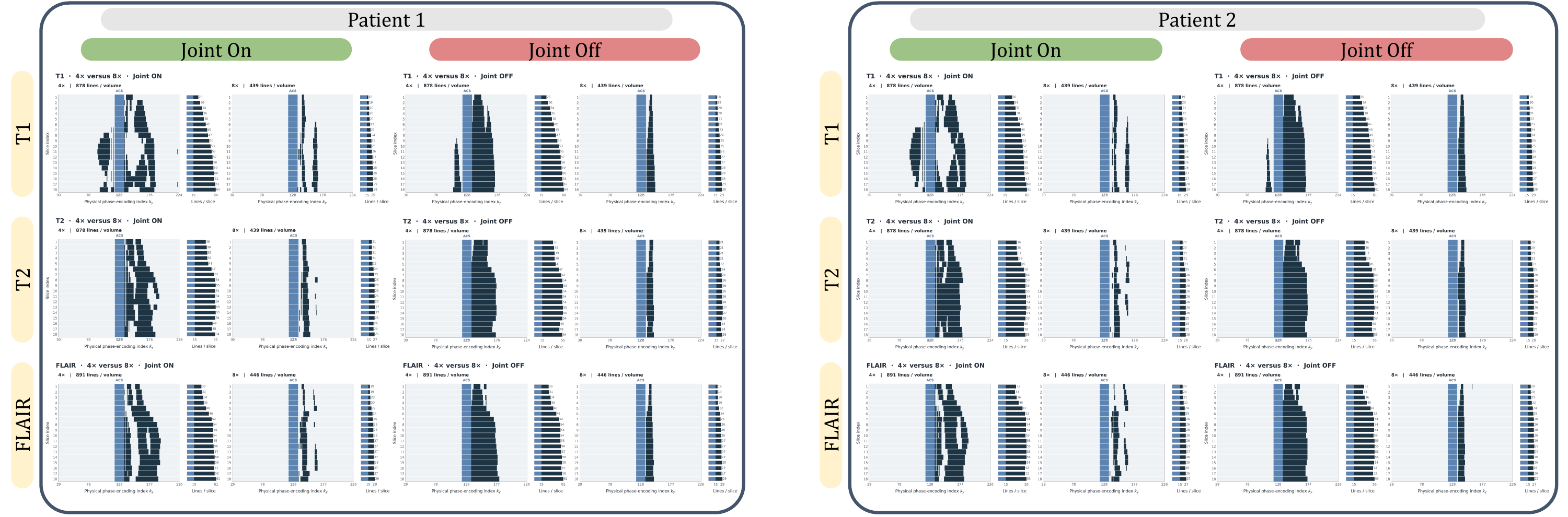}
    \caption{
    Visualization of learned sampling masks for representative M4Raw subjects across different contrasts and acceleration factors.
    }
    \label{fig:mask_ablation}
\end{figure*}

\paragraph{Implementation Details.}
Our framework employs a full-volume ACS-conditioned 3D CNN with implicit coordinate-based line scoring and learned joint slice--frequency refinement.
Reconstruction uses ACS-based coil-sensitivity estimation and 12 synchronous cross-slice reconstruction cascades, each enforcing hard multi-coil k-space data consistency.
All networks are trained from scratch for 100 epochs using AdamW with an initial learning rate of $10^{-4}$, weight decay of $10^{-4}$, linear warm-up, and cosine learning-rate decay.
Training uses two NVIDIA A100 80GB GPUs, processing one full volume per GPU for Knee and 12 volumes per GPU for M4Raw.
We report peak signal-to-noise ratio (PSNR) and structural similarity index (SSIM) for each MRI contrast and acceleration factor.
Following PromptMR+, volume-level PSNR is computed from the reconstruction mean squared error, while SSIM is averaged across slices, using the reference intensity range.

\subsection{Quantitative Comparisons}
\label{sec:quantitative}
For fastMRI Knee, we compare our framework against Inter-Slice~\cite{HimMev_InterSlice_MICCAISAT2026}, U-Net~\cite{ronneberger2015u}, i-RIM~\cite{putzky2019invert}, E2E-VarNet~\cite{sriram2020end}, HUMUS-Net~\cite{fabian2022humus}, PromptMR~\cite{xin2023fill}, and PromptMR+~\cite{xin2024rethinking}.
For M4Raw Brain, we compare against E2E-VarNet~\cite{sriram2020end}, MoDL~\cite{aggarwal2018modl}, and PromptMR~\cite{xin2023fill}, with the baseline models adapted to the M4Raw dataset.

\paragraph{FastMRI Knee.}
Table~\ref{tab:knee_results} presents quantitative comparisons on the fastMRI multi-coil knee dataset at $4\times$ and $8\times$ acceleration, reporting PSNR and SSIM separately for PD and PDFS acquisitions.
We compare the proposed volume-adaptive acquisition and reconstruction framework with established reconstruction methods using their corresponding sampling protocols.
Unlike conventional approaches that employ predefined sampling patterns, our framework jointly determines slice-specific acquisition budgets and sampling locations based on the anatomical information extracted from each MRI volume.
Our method achieves superior reconstruction performance compared with existing approaches across different MRI contrasts and acceleration factors, demonstrating the effectiveness of jointly optimizing volume-adaptive acquisition and reconstruction.
Notably, our method at $8\times$ matches PromptMR+ at $4\times$ in PD PSNR (40.85 dB) while achieving higher SSIM (95.44\% vs.\ 95.00\%), demonstrating comparable or better quality with fewer measurements and the potential for faster acquisition.

\paragraph{M4Raw Brain.}
Table~\ref{tab:m4raw_results} presents quantitative comparisons on the M4Raw low-field brain MRI dataset at $4\times$ and $8\times$ acceleration.
We report PSNR and SSIM separately for T1-weighted, T2-weighted, and FLAIR acquisitions, using the first acquisition repetition from the validation set.
All methods are evaluated on the same validation data under the corresponding acquisition protocols.
Our framework achieves superior reconstruction performance compared with existing methods across different MRI contrasts and acceleration factors.

\subsection{Qualitative Comparisons}
\label{sec:qualitative}
We further evaluate reconstruction quality through representative visual comparisons on the fastMRI multi-coil knee dataset.
The comparisons include reconstructed images and corresponding absolute error maps with respect to the GT images, allowing us to assess reconstruction fidelity and anatomical detail preservation.
Figure~\ref{fig:knee_qualitative} presents representative PD and PDFS reconstructions at $8\times$ acceleration on FastMRI Knee.
Compared with our method, existing approaches, including PromptMR+, produce noticeably blurrier reconstructions and fail to recover anatomical details in certain regions.
In contrast, our method preserves anatomical structures and recovers details that are missing or less clearly reconstructed by competing methods.
Additional comparisons are provided in the supplementary material.

\begin{table}[t]
\small
\centering
\caption{Comparison of online sampling-policy inference
time.}
\label{tab:time}
\scriptsize
\setlength{\tabcolsep}{3pt}
\renewcommand{\arraystretch}{1.15}
\resizebox{\columnwidth}{!}{%
\begin{tabular}{llr}
\toprule
Method & Adaptive Sampling& Time (ms/vol.) $\downarrow$ \\
\midrule
Adaptive VarNet~\cite{bakker2022learning}
    & Slice-wise & 253.411 \\
Inter-Slice~\cite{HimMev_InterSlice_MICCAISAT2026}
    & Inter-slice & 5,099.502 \\
SUNO~\cite{gautam2026scan}
    & Scan-adaptive & 1,146.221 \\
PGA-DPS~\cite{kang2026prioraware}
    & Prior-guided & 318.699 \\
\midrule
\textbf{Ours}
    & \textbf{Volume-aware} & \textbf{54.925} \\
\bottomrule
\end{tabular}%
}
\end{table}
\subsection{Analysis of Learned Sampling Patterns}
\label{sec:ablation}
We investigate the adaptive behavior of our acquisition network and contribution of joint slice--frequency refinement through visualizations of learned sampling masks.
Figure~\ref{fig:mask_ablation} presents sampling patterns for representative M4Raw subjects across different MRI contrasts and acceleration factors, with and without joint refinement.
The learned masks vary across patients, demonstrating patient-specific acquisition decisions.
Furthermore, different MRI contrasts, including T1, T2, and FLAIR, produce distinct sampling patterns, reflecting differences in anatomical information and image characteristics.
The sampling patterns also adapt to the requested acceleration factor, with different numbers and locations of acquired lines at $4\times$ and $8\times$.
Notably, the network assigns fewer measurements to the first slices, which typically contain limited anatomical information, while allocating more measurements to central slices with richer anatomical structures.
To examine the effect of joint refinement, we disable the module only during inference while preserving identical volume-wide acquisition budgets.
The joint module modifies the selected phase-encoding locations across neighboring slices, encouraging complementary acquisition patterns that increase sampling diversity across the volume.
These observations illustrate how our method jointly adapts sampling budgets and acquisition locations to patient anatomy, MRI contrast, and acceleration requirements while coordinating measurements across physical slices.

\subsection{Inference Efficiency}
Table~\ref{tab:time} compares the online inference time required to determine acquisition masks
at $8\times$ acceleration on fastMRI Knee.
Our volume-aware sampling policy requires only $54.925ms$ per volume, achieving the lowest reported
latency among the evaluated methods and a $4.6\times$ speedup over Adaptive VarNet.
By jointly determining slice-dependent sampling patterns and acquisition budgets across the entire
volume, our method avoids the computational overhead of repeated sequential acquisition decisions.
This low inference latency suggests the potential for integration into practical MRI acquisition workflows, where sampling decisions must be made efficiently, although prospective validation
on real MRI scanners remains necessary.

\section{Conclusion}
\label{sec:conclusion}
We presented a volume-adaptive MRI framework that jointly learns \emph{where} and \emph{how much} to sample across an MRI volume using initially acquired ACS measurements.
Our method adapts slice-specific acquisition budgets and k-space line locations while coordinating sampling under a volume-wide budget.
Sensitivity-aware synchronous cross-slice reconstruction exploits neighboring-slice information while enforcing hard multi-coil k-space data consistency.
Experiments on fastMRI Knee and M4Raw Brain show effective reconstruction and adaptive sampling across different anatomies, MRI contrasts, and acceleration factors.

\section*{Acknowledgments}
This work was supported by the InnoCORE program of the Ministry of Science and ICT~(26-InnoCORE-01).

{
    \small
    \bibliographystyle{ieeenat_fullname}
    \bibliography{main}
}


\end{document}